\documentclass[a4paper,10pt]{article}
\usepackage{amssymb}
\usepackage{geometry}
\usepackage[usenames,dvipsnames]{color}
\usepackage{slashed}
\usepackage[table]{xcolor}
\usepackage{graphicx}
\usepackage{mathtools}
\usepackage{cite} 

\usepackage{amsthm}
\usepackage{bm}

\theoremstyle{plain}

\theoremstyle{definition}

\theoremstyle{remark}

\begin{document}

\title{Stochastic gradient descent with random learning rate}
\author{Daniele Musso\footnote{daniele.musso@usc.es, d.musso@inovalabs.es, mudaniele@yahoo.com}}
\date{}

\maketitle
\vspace{-20pt}
\begin{center}\it{
Departamento de  F\'\i sica de Part\'\i  culas,\\
Universidade de Santiago de Compostela\\
\vspace{5pt}
Instituto Galego de F\'\i sica de Altas Enerx\'\i as (IGFAE)\\
E-15782 Santiago de Compostela, Spain\\
\vspace{5pt}
and\\
\vspace{5pt}
Inovalabs Digital S.L. (TECHEYE), E-36202 Vigo, Spain}
\end{center}
\vspace{5pt}

\begin{abstract}
We propose to optimize neural networks with a uniformly-distributed random learning rate.
The associated stochastic gradient descent algorithm can be approximated by continuous stochastic equations and analyzed within the Fokker-Planck formalism.
In the small learning rate regime, 
the training process is characterized by an effective temperature which depends on the average learning rate, 
the mini-batch size and the momentum of the optimization algorithm. By comparing the random learning rate protocol with cyclic and constant protocols,
we suggest that the random choice is generically the best strategy in the small learning rate regime,
yielding better regularization without extra computational cost.
We provide supporting evidence through experiments on both shallow, fully-connected 
and deep, convolutional neural networks for image classification on the MNIST and CIFAR10 datasets.
\end{abstract}
\newpage
\tableofcontents

\section{Introduction}

The stochastic gradient descent (SGD) algorithm is a keystone for neural network training and machine learning at large
\cite{robbins1951,10.1007/978-3-7908-2604-3_16}.
Its simplicity, robustness and regularization effect are crucial strengths which make SGD a standard optimization algorithm 
in a wide range of applications, often outperforming more sophisticated approaches.
Besides, SGD allows for statistical mechanics treatments, such as
an approximate description in terms of continuous stochastic processes \cite{DBLP:journals/corr/LiTE15} 
and the Fokker-Planck equation \cite{DBLP:journals/corr/abs-1710-11029}.
Thereby, SGD provides a privileged access to the theoretical description of the learning dynamics in neural networks
\cite{DBLP:journals/corr/Shwartz-ZivT17}.

SGD has a parameter that controls the rate at which the state of the neural network is updated at each optimization 
step. This is the \emph{learning rate}. In the simplest implementations of SGD, the learning rate 
is constant in time. However, one can consider time schedulings where the learning rate varies throughout the optimization process.
For instance, cyclic protocols where the learning rate is modulated by a cosine have been recently proposed
\cite{DBLP:journals/corr/Smith15a,DBLP:journals/corr/LoshchilovH16a}.
Such cyclic protocols can be thought of as interpolating procedures between two limiting situations:
a constant learning rate (in the infinite period limit), and a uniformly random learning rate (in
a sort of zero period limit). This latter constitutes the main focus of the present analysis.

On the practical side, we provide evidence that all the above mentioned learning protocols perform in an equivalent manner 
on the simplest problems and in the small learning rate regime. This is corroborated by numerical experiments on both a fully-connected, bi-layer 
perceptron and on a deep convolutional neural network employed for image classification on the 
Modified National Institute of Standards and Technology (MNIST) dataset \cite{726791}. 
However, as soon as the complexity of the task is increased, the experiments indicate that the random protocol
performs better than the constant protocol and it is as good as the best among the cyclic protocols. 
Thus, the random learning rate strategy proves to be practically equivalent to a cyclic protocol 
where the period has been fine tuned to yield the best results.
We support this with image classification experiments on the Canadian Institute For Advanced Research (CIFAR10) dataset \cite{cifar10}
with a VGG16 deep neural network \cite{2014arXiv1409.1556S}.

On the theoretical side, we apply an approximated Fokker-Planck analysis to describe the cases with random learning rate.
By so doing, we define an effective temperature in terms of the average learning rate, the size of the 
mini-batches and the momentum. Assuming that equilibrium were reached as the result of the learning process,
the effective temperature would control the eventual probability distribution of the state of the trained network. 
Such equilibrium hypothesis is debated in the literature \cite{DBLP:journals/corr/abs-1710-11029}, and should probably be avoided.
Nevertheless, we experimentally observe that the effective temperature offers a useful handle to classify not 
only the result of the training but the training process itself. 
Specifically, the experiments performed on both shallow and deep cases confirm the practical equivalence among different choices 
of the learning parameters (\emph{i.e.} average learning rate, mini-batch size and momentum) as long as they correspond to the same effective temperature. 

The paper is structured as follows. In Section \ref{ran_lea} we describe the SGD algorithm with random learning rate,
and illustrate its approximate analysis in terms of continuous stochastic equations and the Fokker-Planck formalism.
In Section \ref{exp} we describe the models and the numerical experiments performed to check the theoretical expectations.
We conclude in Section \ref{discu} with some final remarks and questions for future work.

\section{Stochastic equations for SGD with random learning rate}
\label{ran_lea}

Throughout the training, the weight vector $\boldsymbol x$ encoding the internal state of the neural network is updated following the step-wise SGD rules
\begin{align}\label{SGD_up1}
 \boldsymbol v_{k+1} &= \mu\, \boldsymbol v_k - \alpha \, \boldsymbol \nabla_{(x)} f_\Gamma(\boldsymbol x_k)\ ,\\ \label{SGD_up2}
 \boldsymbol x_{k+1} &= \boldsymbol x_k + l\, \boldsymbol v_{k+1}\ ,
\end{align}
where the latin sub-indexes label the discretized time, $\boldsymbol v$ is the velocity, $\mu$ is the momentum,
$\boldsymbol \nabla_{(x)}$ is the gradient with respect to the weights $\boldsymbol x$, 
$l$ controls the update rate of $\boldsymbol x$ and $\alpha$ is a random variable uniformly distributed in the interval $[1-\Delta,1+\Delta]$
with $\Delta\leq1$. In \eqref{SGD_up1} we have also defined $f_\Gamma(\boldsymbol x)$, the loss function evaluated on a sampled mini-batch $B_\Gamma$, 
\begin{equation}
 f_\Gamma(\boldsymbol x) = \frac{1}{C} \sum_{j\in B_\Gamma} f_j(\boldsymbol x)\ ,
\end{equation}
where the index $\Gamma$ labels the mini-batches. Each mini-batch is a collection of $C$ input instances.
Note that $l$ would be the usual learning rate if $\alpha$ were constant and 
equal to $1$, corresponding to $\Delta = 0$.

The SGD updating rules \eqref{SGD_up1} and \eqref{SGD_up2} can be rewritten as follows
\begin{align}\label{SGD_up1b}
 \boldsymbol v_{k+1} -\boldsymbol v_k &= -l \left[\frac{1-\mu}{l}\, \boldsymbol v_k + \frac{1}{l}\, \boldsymbol \nabla_{(x)} f(\boldsymbol x_k)\right] 
 + \boldsymbol \xi (\boldsymbol x_k) \ ,\\ \label{SGD_up2b}
 \boldsymbol x_{k+1} - \boldsymbol x_k &= l\, \boldsymbol v_{k+1}\ ,
\end{align}
where $f$ is the loss function on the whole training dataset%
\footnote{
For later convenience, we introduce some useful notation:
\begin{equation}
 f(\boldsymbol x) = \frac{1}{\cal N} \sum_\Gamma f_\Gamma (\boldsymbol x)
 = \frac{1}{{\cal N} C} \sum_\Gamma \sum_{j\in B_\Gamma} f_j(\boldsymbol x)
 =\frac{1}{N} \sum_{i=1}^N f_i(\boldsymbol x)\ ,
\end{equation}
where $N = {\cal N} C$ is the total number of images in the training dataset,
${\cal N}$ is the number of mini-batches and $C$ is the number of images for each mini-batch.
The index $\Gamma$ runs over the mini-batches and $B_\Gamma$ denotes a mini-batch instance.},
and $\boldsymbol \xi(\boldsymbol x)$ is the random vector
\begin{equation}
 \boldsymbol \xi (\boldsymbol x) = \boldsymbol \nabla_{(x)} f (\boldsymbol x) - \alpha\, \boldsymbol \nabla_{(x)} f_\Gamma(\boldsymbol x) \ .
\end{equation}
In order to have $\langle \boldsymbol \xi (\boldsymbol x) \rangle = 0$, it is necessary to have $\langle \alpha \rangle = 1$;%
\footnote{We thank Takashi Mori for discussions on this point.} note also that the expectation of the noise $\boldsymbol \xi( \boldsymbol x)$ 
is to be taken both with respect to the 
random mini-batch distribution and to the distribution for $\alpha$.
Observing \eqref{SGD_up1b}, recall that the momentum $\mu \in [0,1]$, so it leads to a positive friction coefficient $\gamma = \frac{1 - \mu}{l}$.
The covariance matrix of the random vector $\boldsymbol \xi (\boldsymbol x)$ is approximately
\begin{equation}\label{cicova}
 \langle \boldsymbol \xi(\boldsymbol x) \boldsymbol \xi^T(\boldsymbol y) \rangle 
 \approx \frac{1}{C}\ \hat D(\boldsymbol x)\ \delta^{(n)} (\boldsymbol x-\boldsymbol y)\ ,
\end{equation}
where $C$ is the mini-batch size and the approximate diffusion matrix $\hat D(\boldsymbol x)$ is introduced in \eqref{diffu4}.
We denoted with $n$ the total number of weights in the network, \emph{i.e.} the dimensionality of $\boldsymbol x$.
The angular brackets in \eqref{cicova} indicate again averaging both with respect to the mini-batch sampling and on the random parameter 
$\alpha \in [1-\Delta, 1+\Delta]$. In the limit $\Delta\rightarrow 0$, expression \eqref{cicova} becomes exact, see Appendix \ref{difma} for details.

We consider the simplifying hypothesis of an isotropic diffusion matrix
\begin{equation}\label{iso}
 \hat D(\boldsymbol x) = D\ \hat I_{n\times n}\ ,
\end{equation}
where $D$ is a number and $\hat I_{n\times n}$ is the $n$-dimensional identity matrix.
It must be stressed that the isotropic hypothesis \eqref{iso} is \emph{not} an assumption to be taken lightly.
\cite{DBLP:journals/corr/abs-1710-11029} argues that the generic case entails non-isotropic 
diffusion matrices, especially in dealing with deep neural networks.%
\footnote{See also \cite{musso2020partial} for a recent study of weight-space anisotropy in deep neural networks.}
Despite this, we show that 
the isotropic assumption does not spoil the practical utility of the present analysis.

Relying on \cite{DBLP:journals/corr/LiTE15}, the stochastic process \eqref{SGD_up1b} and \eqref{SGD_up2b}
can be approximated by the continuous stochastic system of equations
\begin{align}\label{sto_sys_1}
 \frac{d \boldsymbol V}{dt} &= -\frac{1-\mu}{l}\, \boldsymbol V 
 - \frac{1}{l} \boldsymbol \nabla_{(X)} f(\boldsymbol X) + \sqrt{\frac{D}{C}}\, \frac{d \boldsymbol W}{dt}\ ,\\ \label{sto_sys_2}
 \frac{d \boldsymbol X}{dt} &= \boldsymbol V\ ,
\end{align}
where $\boldsymbol W$ represents a Wiener process corresponding to noise. A couple of comments are in order.
The stochastic system \eqref{sto_sys_1} and \eqref{sto_sys_2} approximates the discrete stochastic process
\eqref{SGD_up1b} and \eqref{SGD_up2b} in an order $1$ weak distributional sense, as proved in \cite{DBLP:journals/corr/LiTE15},
where the order of the approximation refers to the time step $dt = l$. Higher order approximations, $l^n$, are possible if one considers
the Euler-Maruyama discretization and matches the moments \cite{DBLP:journals/corr/LiTE15}. We therefore expect the present analysis to work in the small 
learning rate limit $l\ll 1$.

The random variable $\boldsymbol W$ encodes gradient noise and is assumed to be normally distributed, such hypothesis is common in the literature. 
Nonetheless, this is another assumption not to be taken lightly. Non-trivial correlations deviating from Gaussianity are a typical feature in machine learning.%
\footnote{For instance, the weights of a neural network are generically non-Gaussian, especially after being trained.}
Assuming Gaussian noise is possibly connected to a mean field approximation,
whose validity should be nevertheless taken with a critical attitude.

\subsection{Fokker-Planck analysis}
\label{FP}

Consider the phase space probability density $\rho(\boldsymbol V, \boldsymbol X, t)$
where the random variables $\boldsymbol X$ and $\boldsymbol V$ represent a vector position and a velocity respectively.
Because of the conservation of the total probability, $\rho$ obeys the generalized continuity equation
\begin{equation}\label{con}
 \frac{\partial}{\partial t} \rho(\boldsymbol V, \boldsymbol X, t)
 + \boldsymbol \nabla_{(X)} \cdot \left[ \frac{d \boldsymbol X}{dt}\, \rho(\boldsymbol V, \boldsymbol X, t) \right]
 + \boldsymbol \nabla_{(V)} \cdot \left[ \frac{d \boldsymbol V}{dt}\, \rho(\boldsymbol V, \boldsymbol X, t) \right] = 0\ .
\end{equation}
We indicate with $P(\boldsymbol V, \boldsymbol X, t) = \langle \rho (\boldsymbol V, \boldsymbol X, t) \rangle$
the average of the probability density over many realization of the batch noise and the random learning parameter $\alpha$.
By a standard analysis%
\footnote{See Appendix \ref{fokpla} for the explicit computation.}, 
$P$ obeys a Fokker-Planck equation which, using \eqref{sto_sys_1} and \eqref{sto_sys_2},
takes the following explicit form:
\begin{equation}\label{ffpp}
 \frac{\partial P}{\partial t}
 = -\boldsymbol \nabla_{(X)} \cdot \left(\boldsymbol V P\right)
 -\boldsymbol \nabla_{(V)} \cdot \left[\left(\frac{\mu - 1}{l}\boldsymbol V - \frac{1}{l} \boldsymbol \nabla_{(X)} f \right) P\right]
 + \frac{D}{2C}\ \boldsymbol \nabla_{(V)}^2 P\ .
\end{equation}

If one assumes to be at equilibrium, then $\frac{\partial P}{\partial t} = 0$.
The Fokker-Planck equation reduces to
\begin{equation}\label{eq_FP}
 -\boldsymbol \nabla_{(X)} \cdot \left(\boldsymbol \nabla_{(V)}H\ P\right)
 -\boldsymbol \nabla_{(V)} \cdot \left\{\left[\frac{\mu - 1}{l}\boldsymbol \nabla_{(V)}H - \frac{1}{l} \boldsymbol \nabla_{(X)} f \right] P\right\}
 + \frac{D}{2C}\ \boldsymbol \nabla_{(V)}^2 P = 0\ ,
\end{equation}
where we have introduced the Hamiltonian
\begin{equation}
 H (\boldsymbol V, \boldsymbol X) = \frac{1}{2}\ \boldsymbol V \cdot \boldsymbol V + \frac{1}{l} f(\boldsymbol X)\ ,
\end{equation}
and used $\boldsymbol V = \boldsymbol \nabla_{V} H$.
This Hamiltonian generates the conservative and deterministic part of the dynamical equation \eqref{sto_sys_1}.

Consider the following manipulations
\begin{align}\label{man}
 &-\boldsymbol \nabla_{(X)} \cdot \left(\boldsymbol \nabla_{(V)}H\ P\right)
 +\frac{1}{l} \boldsymbol \nabla_{(V)} \cdot \left(\boldsymbol \nabla_{(X)} f\ P\right)\\ \nonumber
 =&- \left(\boldsymbol \nabla_{(V)}H\right) \cdot \left(\boldsymbol \nabla_{(X)}\ P\right)
 + \left(\boldsymbol \nabla_{(X)} H\right) \cdot \left( \boldsymbol \nabla_{(V)} P\right) = 0\ ,
\end{align}
where we have used $\boldsymbol \nabla_{(V)} \cdot \boldsymbol \nabla_{(X)} H = 0$ and assumed $P = P(H)$.
Using \eqref{man}, the equilibrium Fokker-Planck equation \eqref{eq_FP} reduces to
\begin{equation}
 \boldsymbol \nabla_{(V)} \cdot \left\{ \boldsymbol V \left[\frac{\mu - 1}{l} P
 - \frac{D}{2C}\  P' \right] \right\}= 0\ ,
\end{equation}
and it is solved by
\begin{equation}
 P = \frac{1}{Z} \exp\left[-\frac{2 C (1-\mu)}{l\, D}\  H\right]\ ,
\end{equation}
where $Z$ is a normalizing factor interpreted as a partition function.
In fact, one can define the ``free energy'' potential
\begin{equation}
  F = - T \log P = H - ST\ ,
\end{equation}
and have an effective thermodynamic interpretation of the equilibrium state
where $S = -\log Z$ represents an entropy. We have thus identified an effective temperature
\begin{equation}\label{eff_tem}
 T = \frac{l\, D}{2C (1-\mu)}\ ,
\end{equation}
which depends on the (average) learning rate $l$, the mini-batch size $C$ and the momentum $\mu$.

The effective thermodynamic analysis in the presence of a random learning rate 
works in a similar way as in the case with constant learning rate.

\section{Experiments}
\label{exp}

We consider three neural networks:
\begin{itemize}
 \item {\bfseries Model 1}, defined in \ref{model1} is a bi-layer perceptron.
 \item {\bfseries Model 2}, defined in \ref{model2} is a deep convolutional network with two covnolutional blocks 
 followed by a fully connected classification layer. The convolutional kernel size is $3\times 3$ throughout the 
 entire network.
 \item {\bfseries Model 3}, is the sixteen-layer VGG deep convolutional network introduced in \cite{2014arXiv1409.1556S}.%
 \footnote{We have used the vgg16 Pytorch implementation with pretrained weights on the ImageNet database \cite{deng2009imagenet}.}
\end{itemize}

\noindent
We compare different learning protocols, specifically:
\begin{enumerate}
 \item The random learning rate described above.
 \item A constant learning rate protocol.
 \item Three cyclic, cosinusoidal protocols with periods $P\in\{6,18,30\}$ defined by 
 \begin{equation}
  l_\tau = l \left[1 + \cos\left(\frac{\pi \tau}{P}\right)\right]\ ,
 \end{equation}
where $\tau$ counts the number of completed training epochs.%
\footnote{The symbol $P$ has two meanings throughout the paper, it denotes the phase space probability density $P(\boldsymbol V,\boldsymbol X,t)$ in the 
Fokker-Planck analysis of Subsection \ref{FP} and the period of the cyclic schedule everywhere else.}
\end{enumerate}

\subsection{Equivalence of random, cyclic and constant learning rates on MNIST}
\label{MNIST}
\begin{figure}
 \begin{center}
  \includegraphics[width=0.49\textwidth]{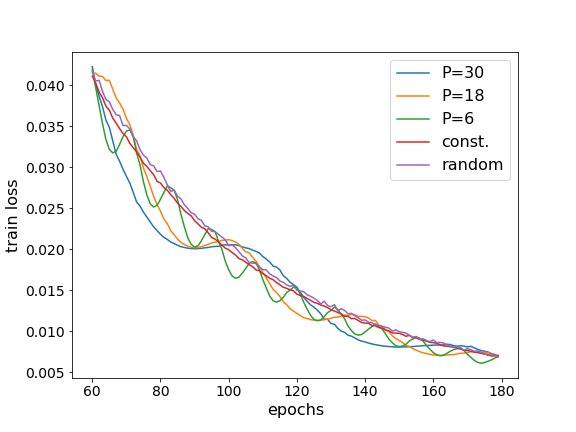}
  \includegraphics[width=0.49\textwidth]{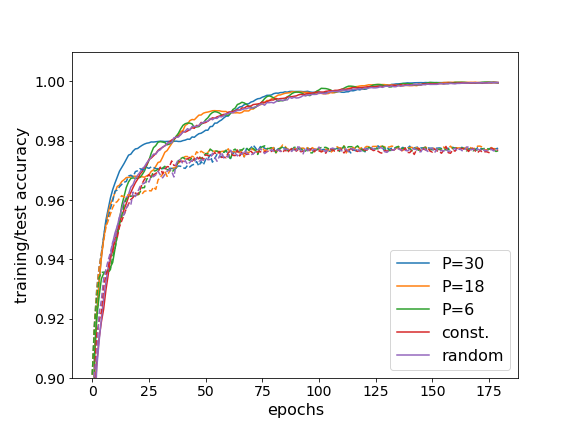}
 \end{center}
 \caption{Comparison of different learning rate protocols in a bi-layer perceptron (Model 1) on MNIST with $l=5\; 10^{-3}$:
 constant learning rate; uniformly distributed random learning rate; cyclic learning rates with 
 period of $6, 18, 30$ epochs, respectively. The plots depict the training loss (left) and the training/test 
 accuracy (right). Results have been obtained using mini-batch size $C=256$, momentum $\mu=0.9$, 
 Nesterov acceleration and no weight decay regularization.  
 \label{com_sha}
 }
\end{figure}

\begin{figure}
 \begin{center}
  \includegraphics[width=0.49\textwidth]{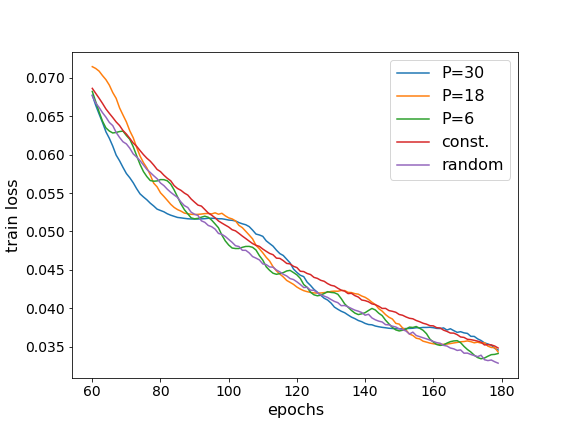}
  \includegraphics[width=0.49\textwidth]{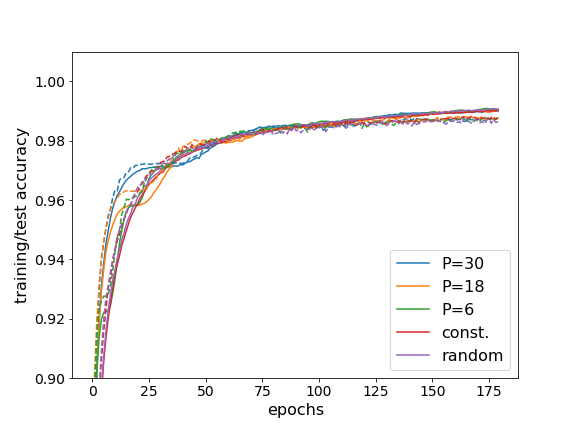}
 \end{center}
 \caption{Comparison of different learning rate protocols in a convolutional deep network (Model 2) on MNIST with $l=5\; 10^{-5}$:
 constant learning rate; uniformly distributed random learning rate; cyclic learning rates with 
 periods of $6, 18, 30$ epochs. The plots depict the training loss (left) and the training/test 
 accuracy (right). Results have been obtained using mini-batch size $C=256$, momentum $\mu=0.9$, 
 Nesterov acceleration and weight decay $w=10^{-3}$.  
 }
 \label{com_dee}
\end{figure}

We first consider an individual training example on MNIST for each protocol and show them in Figure \ref{com_sha} and Figure \ref{com_dee}.
In each figure, all the training instances depicted correspond to the same mean value $l$ of the learning rate.
This means that, during all training instances, the weights are on average updated according to $- l\, \boldsymbol \nabla_{(x)} f_\Gamma (\boldsymbol x)$ 
for each sampled mini-batch $\Gamma$. The average $l$ coincides with the learning rate value of the constant protocol.

Figure \ref{com_sha} and Figure \ref{com_dee} show that all the tested learning protocols 
proved to be equivalent to practical purposes on MNIST. More specifically, the trainings have all reached 
equivalent levels of training loss, train accuracy and test accuracy. Thus, the methods entails here
no advantage as the minimization process is concerned, nor they differ in the regularization effect.
In fact, the training/test accuracy results on MNIST show no difference concerning the generalization error.
These observations apply to both the shallow and the deep case in the small learning rate 
regime. Nonetheless, we anticipate that the situation changes when the complexity of the classification task is increased.

\begin{figure}
 \begin{center}
  \includegraphics[width=0.49\textwidth]{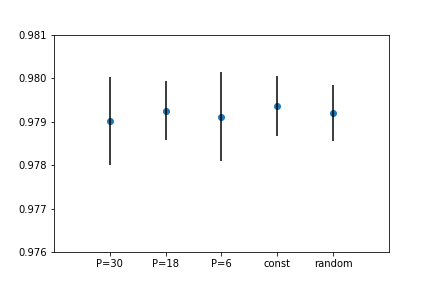}
  \includegraphics[width=0.49\textwidth]{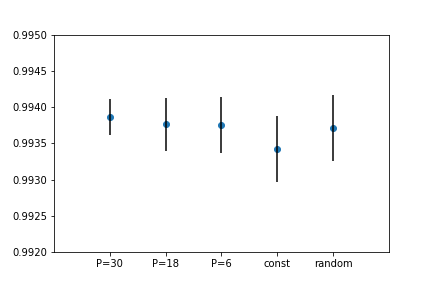}
 \end{center}
 \caption{Left plot: Comparison between the best test-accuracy values obtained on the MNIST dataset with different learning rate protocols
 in a bi-layer perceptron (Model 1) with no weight-decay regularization.
 Right plot: Same experiments but using a deep convolutional network (Model 2)
 with weight-decay regularization $w=10^{-3}$. All experiments refer to $180$ training epochs with learning rate $l=0.005$, 
 momentum $\mu=0.9$ and mini-batch size $C=256$. The error bars represent the standard deviation.
 }
 \label{syste}
\end{figure}

We then perform a more systematic analysis on MNIST, collecting the best test accuracy 
reached by both Model 1 and Model 2 (see Appendix \ref{Ann}) over a series of training instances with different pseudo-random seeds.
The results are reported in Figure \ref{syste}. 
\begin{figure}
 \begin{center}
  \includegraphics[width=0.49\textwidth]{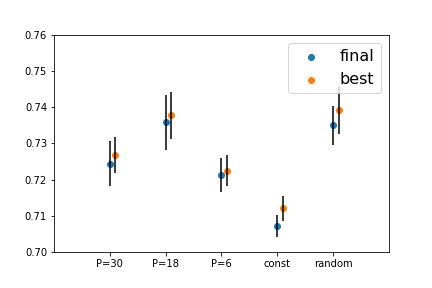}
  \includegraphics[width=0.49\textwidth]{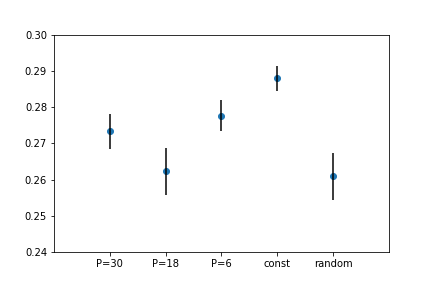}
 \end{center}
 \caption{Left plot: Comparison between the final and best test-accuracy values obtained on the CIFAR10 dataset with the VGG16 architecture (Model 3)
 with different learning protocols; the error bars represent the standard deviation. The random protocol performs slightly better than the best among the cyclic 
 protocols. Right plot: generalization error from the difference between the training (in sample) and test (out of sample) accuracy 
 for VGG16 on CIFAR10. The error bars descend from propagation of the standard deviations of the original data. This plot roughly reproduces 
 the flipped trend of the left plot, meaning that the difference in performance of the various protocols is essentially due to a different 
 regularization power. All experiments refers to $180$ training epochs with a learning rate $lr=0.0005$, momentum $\mu=0.9$, mini-batch size $C=256$ 
 and weight-decay regularization $w=10^{-3}$.
 }
 \label{cifra}
\end{figure}

Despite referring to different learning protocols, all the training instances compared with each other 
refer to the same value of the effective temperature \eqref{eff_tem}. In particular, the constant and random protocols 
are approximated by analogous continuous stochastic processes \eqref{sto_sys_1}, \eqref{sto_sys_2}.%
\footnote{There are differences at the level of the diffusion matrix \eqref{diffu4} which are not important 
for the present discussion.}
A similar, though weaker, statement is true for the cyclic protocols as well, admitting that the continuous 
stochastic process is approximating the actual stochastic process only after averaging over many periods.

The higher order analysis performed in \cite{DBLP:journals/corr/LiTE15} offers a hint to explain why one is expected 
to see deviations as soon as the learning parameter $l$ is increased. In our case, we do not expect the practical equivalence 
of the various learning protocols on MNIST to hold true at higher values of $l$. To have a finer approximation to the stochastic 
process \eqref{SGD_up1}, \eqref{SGD_up2}, the continuous system \eqref{sto_sys_1}, \eqref{sto_sys_2} needs higher order 
terms in the learning rate $l$. Such terms would also affect the subsequent Fokker-Planck analysis. As a consequence,
the role of $l$ in setting the properties of the probability distribution of the network states (and thus the effective temperature $T$) 
can be modified once higher values of $l$ are considered.

More specifically, \cite{DBLP:journals/corr/LiTE15} shows that the first non-trivial correction to the right hand side of \eqref{SGD_up1} 
at higher order in $l$ is%
\footnote{\cite{DBLP:journals/corr/LiTE15} considers a case without momentum.}
\begin{equation}\label{corre}
 - \frac{l}{4}\ \boldsymbol \nabla_{(X)} f(\boldsymbol X) \cdot \boldsymbol \nabla_{(X)} f(\boldsymbol X)\ .
\end{equation}
For higher dimensional weight spaces the correction term \eqref{corre} is expected to be bigger, so it is easier to exit the small rate regime.
Even though a proper quantitative definition of the small rate regime would require knowledge of the actual gradient 
of the loss function throughout the dataset, we expect it to correspond to a stricter 
requirement in the case of deep architectures. We actually observed this fact directly: indeed, in order to show 
the practical equivalence among the different protocols on MNIST, the data collected with the deep network (Figure \ref{com_dee})
required a value for $l$ which is two orders of magnitude smaller than those collected with the bi-layer 
perceptron (Figure \ref{com_sha}).

\subsection{Enhanced regularization on CIFAR10}
\label{CIFAR10}

The results obtained on the CIFAR10 dataset have important differences from those obtained on MNIST.
Mainly, the different learning protocols are here \emph{not} practically equivalent among each others and the random protocol
yields results which are as good as the best obtained with its cyclic competitors; the constant protocol proves instead to be the worst (see Figure \ref{cifra}).
The fact that the random learning protocols performs as the best among the cyclic protocols suggests that it is equivalent to
a cyclic protocol whose period is fine-tuned to be optimal. Thereby, the random strategy appears to reach optimality without requiring 
the extra parameter associated to the cyclic period $P$ and its optimization.%
\footnote{Such observation is particularly relevant because the cyclic protocol itself has been introduced as a ``universal 
approach'' which reduces the need to search for optimal training hyperparameters \cite{DBLP:journals/corr/Smith15a}.}

CIFAR10 leads to a more demanding classification task with respect to MNIST,
this is one reason why we observe more sensitivity to regularization. In words, the model 
finds an increased difficulty in getting rid of irrelevant information during training (\emph{i.e.} compression).
Another reason comes from the architecture employed, \emph{i.e.} VGG16, which is deeper and more complex than the models adopted 
for MNIST (Model 1 and Model 2). 
To verify that the different values for the test accuracy among the learning protocols is actually due to regularization effects, we plot 
the generalization error, namely the difference between the training and the test accuracy, see right plot in Figure \ref{cifra}; this provides a quantitative 
estimate of the overfitting. Since the generalization error data points reproduce approximately the same -flipped- behavior as the 
test accuracies, the variations in the latter are due to a different level of regularization.

Arguably, the results obtained on CIFAR10 with VGG16 refer to a generic family of classification tasks that are intrinsically 
more complicated that handwritten characters and which typically require the use of deep neural architectures. 
In this sense, the random protocol is likely to be an optimal strategy for a wide class of learning tasks.

\subsection{Effective temperature to characterize different trainings}
\label{tempera}

The effective temperature \eqref{eff_tem} provides a good criterion 
to characterize the training process and after training performance of the neural networks trained with random learning rate protocol.
We remind the reader and stress that the effective temperature 
is in principle just an equilibrium property, and that equilibrium is a delicate assumption, to say the least.
Nevertheless, the utility of the effective temperature to classify equivalent (in practice) training is 
apparent from the experiments. This could possibly hint that the conditions through training and 
at the end of it are those of a quasi-equilibrium.%

\begin{figure}
 \begin{center}
  \includegraphics[width=0.49\textwidth]{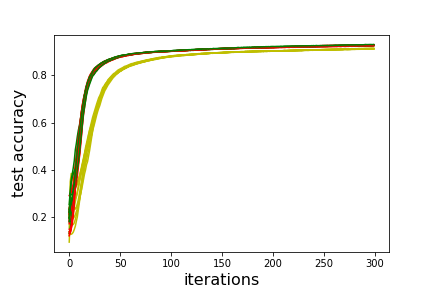}
  \includegraphics[width=0.49\textwidth]{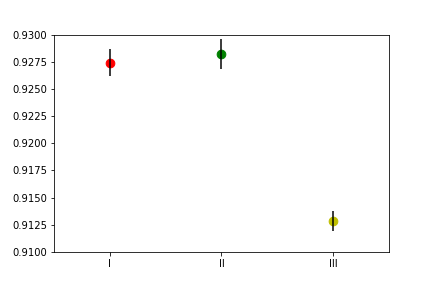}
 \end{center}
 \caption{Equal temperature trainings of a convolutional network (Model 1) on MNIST with the random learning rate protocol:
 red lines (experiment I) have been obtained with a mini-batch size $C=60$ and learning rate $l=2.\; 10^{-4}$
 while the green lines (experiment II) correspond to $C=30$ and $l=1.\; 10^{-4}$.  
 The pale yellowish lines (experiment III) represent training instances with $C=60$ and $l=1.\; 10^{-4}$,
 which thus correspond to a different temperature and are reported just as a counter-check.
 All trainings have been performed with zero momentum. The right plot 
 depicts the values and standard deviations of the test accuracy obtained in the three experiments.
 }
 \label{dif_tem_sha}
\end{figure}

\begin{figure}
 \begin{center}
  \includegraphics[width=0.49\textwidth]{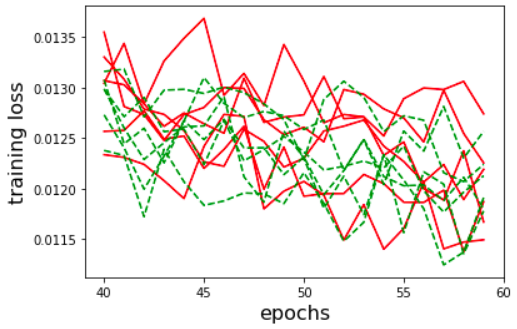}
  \includegraphics[width=0.49\textwidth]{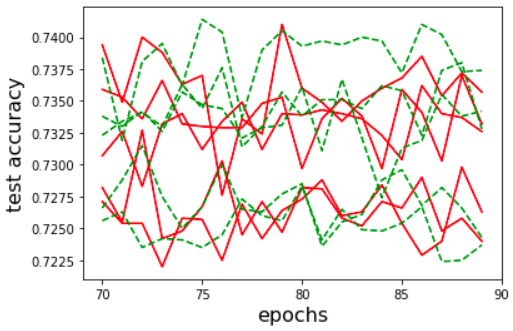}
 \end{center}
 \caption{{\bfseries Left}: Equal temperature trainings of a deep convolutional network (Model 2) on MNIST;
 red solid lines have been obtained with a mini-batch size $C=60$ and momentum $\mu = 0.75$
 while the green dashed lines correspond to $C=30$ and $\mu = 0.5$.  
 All trainings have been performed with a random learning rate uniformly distributed 
 in the interval $[0,0.005]$ and Nesterov acceleration.
 {\bfseries Right}: Equal temperature trainings of VGG16 (Model 3) on CIFAR10;
 red solid lines have been obtained with a mini-batch size $C=128$ and average learning rate $l = 0.00025$,
 their final mean and standard deviation are $0.73036 \pm 0.00444$.
 The green dashed lines correspond to $C=256$ and $l = 0.0005$, their final mean and standard deviation are $0.73002 \pm 0.00526$\ .  
 All trainings have been performed with momentum $\mu = 0.9$ and Nesterov acceleration.
 }
 \label{dif_tem_dee}
\end{figure}

To show the practical utility of the effective temperature, we consider distinct trainings 
associated with different values of the training parameters (the average learning rate $l$, the mini-batch size $C$
and the momentum $\mu$) but corresponding to the same value of the effective temperature \eqref{eff_tem}.
The data collected show that $T$ is the relevant quantity to consider in order to classify the 
performance of different trainings. See Figures \ref{dif_tem_sha} and \ref{dif_tem_dee}.
Note that these results refer to both MNIST and CIFAR10 and to the three neural network architectures,
that is, the bi-layer perceptron (Model 1), the deep convolutional network (Model 2) and the 
deeper convolutional network VGG16 (Model 3).

\section{Related work and future directions}

The present study fits in the overarching field of statistical mechanics applications to machine learning
\cite{doi:10.1146/annurev-conmatphys-031119-050745,Carleo_2019}. In particular it focuses on the effects of 
an additional and controllable randomness on the reduction of the generalization error. Related studies 
about the role of noise over regularization are \cite{Bishop94trainingwith,6796981,article,ali2020implicit}.
More specifically, the effects of direct noise injection and its regularizing power have been addressed in
\cite{SIETSMA199167,neelakantan2015adding}. This connects to a wider question concerning the relation 
between stability and generalization \cite{bousquet2002stability}.

The effective temperature introduced in Subsection \ref{tempera} calls for a more systematic study of the 
underlying statistical ensemble. Interesting and useful analyses to progress in this direction are described in
\cite{PhysRevA.45.6056,marcaccioli2019maximum,marcaccioli2020correspondence,pushkarov2019training}.

The random learning parameter considered in the present study has a direct relation with quantum inspired Hamiltonian Monte Carlo 
algorithms where the mass is randomized. Such correspondence is elucidated in \cite{liu2019quantuminspired}
which offers an alternative application of similar ideas based on an extra source of randomness to the sampling problem.

One further and generic open question emerges from the utility of the effective temperature even
outside of the assumptions needed for its derivation. Specifically, the isotropy of the diffusion
matrix, the equilibrium condition and the Gaussianity of the gradient noise. The main future
direction is to directly tackle such assumptions in order to possibly relax them and increase the
reach and precision of the theoretical control over the SGD algorithm and its generalizations.

\section{Discussion}
\label{discu}

We considered a stochastic gradient descent algorithm with a random learning rate uniformly distributed in an interval.
We characterized this training protocol both theoretically and experimentally, comparing it against alternative protocols
where the learning rate is either constant or periodically modulated.
We focused on the small learning rate regime, where the practical interest is higher and the theoretical modeling easier.
We showed that the training and its performance can be compactly characterized by an effective temperature which is 
directly related to the training hyper-parameters \eqref{eff_tem}.

The effective temperature is derived by means of an approximated and analytic treatment of the stochastic process 
associated to the SGD optimization algorithm. The theoretical analysis applies standard statistical physics methods 
(such as the stochastic equations for Wiener processes and the Fokker-Planck formalism) to the stochastic process
associated to SGD with random learning rate. The source of noise is here twofold, from the mini-batch sampling 
and from the learning rate sampling, respectively. Nevertheless, the SGD with random learning rate admits an approximated 
theoretical treatment analogous to the case of constant rate.

We provided experimental evidence that all the tested learning protocols are equivalent in performance
on MNIST in the small learning rate regime. The results on CIFAR10 show instead a more complicated behavior where
the different protocols perform differently. On the practical level, the experiments on CIFAR10 suggest that the random protocol is the best candidate among 
its competitors, proving to be equivalent to a cyclic protocol whose periodicity is fine-tuned to optimality.
This statement refers to cases where the learning parameter is sufficiently small and is expected to apply 
to a wide family of tasks addressed by means of deep neural architectures.

\section{Acknowledgements}

I would like to thank Alnur Ali, Amparo Alonso Betanzos, Nahuel Arrieta, Alberto Casal, Eva Cernadas Garc\'ia, Manuel Fern\'andez Delgado, 
Stefano Goria, Ziming Liu, Noelia S\'anchez-Maro\~no, Andrea Mezzalira, Takashi Mori, Giorgio Musso, Natalia Nogueira and Maurice Weiler for interesting conversations, 
useful suggestions and support. A special thank to Hern\'an Serrano for his guidance in programming.\\

Part of the computing power needed for the project has been provided by the TechEye project at Inovalabs Digital S.L.

\bibliography{RLR_v4} 
\bibliographystyle{ieeetr}

\appendix

\section{Diffusion matrix}
\label{difma}

We consider the random vector
\begin{equation}
 \boldsymbol \xi (\boldsymbol x) = \boldsymbol \nabla_{(x)} f (\boldsymbol x) - \alpha\, \boldsymbol \nabla_{(x)} f_\Gamma(\boldsymbol x)\ ,
\end{equation}
where the source of randomness is twofold: $\alpha$ is a random variable uniformly distributed in $[1-\Delta,1+\Delta]$ and
the mini-batch $B_\Gamma$ is randomly sampled from the dataset. The loss associated to the mini-batch $B_\Gamma$ is
\begin{equation}
 f_{\Gamma}(\boldsymbol x) = \frac{1}{C} \sum_{j\in B_\Gamma} f_j(\boldsymbol x)\ ,
\end{equation}
where the index $j$ runs over the elements of the mini-batch $B_\Gamma$, $C$ is the mini-batch size
and $\Gamma = \{1,..., {\cal N}\}$ where ${\cal N}$ is the total number of mini-batches.
The covariance matrix of $\boldsymbol \xi (\boldsymbol x)$ is given by
\begin{equation}\label{diffu1}
 \begin{split}
 \Sigma(\boldsymbol x) =& \frac{1}{2 \Delta \cal N} 
 \int_{1-\Delta}^{1+\Delta} d\alpha \sum_{\Gamma} \\
 &\left[\boldsymbol \nabla_{(x)} f(\boldsymbol x) - \frac{\alpha}{C}\sum_{j\in B_\Gamma}\boldsymbol \nabla_{(x)} f_j(\boldsymbol x) \right]
 \left[\boldsymbol \nabla_{(x)} f(\boldsymbol x) - \frac{\alpha}{C}\sum_{k\in B_\Gamma} \boldsymbol \nabla_{(x)} f_k(\boldsymbol x) \right]^T \ .
 \end{split}
\end{equation}
Using
\begin{align}
 \langle \alpha \rangle_\alpha &= \frac{1}{2\Delta} \int_{1-\Delta}^{1+\Delta} d\alpha\, \alpha = 1\ ,\\
 \langle \alpha^2 \rangle_\alpha &= \frac{1}{2\Delta} \int_{1-\Delta}^{1+\Delta} d\alpha\, \alpha^2 = 1 + \frac{\Delta}{3}\ ,
\end{align}
and
\begin{align}
 \langle f_\Gamma(\boldsymbol x) \rangle_\Gamma &= \frac{1}{\cal N} \sum_\Gamma f_\Gamma(\boldsymbol x) = f(\boldsymbol x)\ ,
\end{align}
we can rewrite \eqref{diffu1} as follows:
\begin{equation}
 \Sigma(\boldsymbol x) =
 \frac{1}{{\cal N} C^2} 
 \sum_{\Gamma} \sum_{j\in B_\Gamma} \sum_{k\in B_\Gamma}
 \left(\frac{3+\Delta}{3}\, \boldsymbol \nabla_{(x)} f_j(\boldsymbol x) \boldsymbol \nabla_{(x)} f_k(\boldsymbol x)^T
 - \boldsymbol \nabla_{(x)} f(\boldsymbol x)\boldsymbol \nabla_{(x)} f(\boldsymbol x)^T\right) \ .
\end{equation}
We can split the diagonal and off-diagonal contributions inserting a suitably written $1$, namely
\begin{equation}\label{diffu4}
 \begin{split}
 \Sigma(\boldsymbol x) =& \frac{1}{{\cal N} C^2} 
  \sum_{\Gamma} \sum_{j\in B_\Gamma} \sum_{k\in B_\Gamma}
  \left[\delta_{jk} + \left(1-\delta_{jk}\right)\right] \cdot \\
  &\qquad \qquad \left(\frac{3+\Delta}{3}\, \boldsymbol \nabla_{(x)} f_j(\boldsymbol x) \boldsymbol \nabla_{(x)} f_k(\boldsymbol x)^T
  - \boldsymbol \nabla_{(x)} f(\boldsymbol x)\boldsymbol \nabla_{(x)} f(\boldsymbol x)^T\right) \\
  \equiv& \frac{1}{C}\hat D(\boldsymbol x) + d_\Delta(\boldsymbol x)\ ,
 \end{split}
\end{equation}
where 
\begin{align}
 \hat D(\boldsymbol x) &= \frac{1}{{\cal N} C} 
  \sum_{\Gamma} \sum_{j\in B_\Gamma} 
  \left(\frac{3+\Delta}{3}\, \boldsymbol \nabla_{(x)} f_j(\boldsymbol x) \boldsymbol \nabla_{(x)} f_j(\boldsymbol x)^T
  - \boldsymbol \nabla_{(x)} f(\boldsymbol x)\boldsymbol \nabla_{(x)} f(\boldsymbol x)^T\right)\ , 
\end{align}
comes from the diagonal piece proportional to $\delta_{jk}$,
while $d_\Delta(\boldsymbol x)$ comes from the off-diagonal piece proportional to $\left(1-\delta_{jk}\right)$.
The off-diagonal contribution vanishes for $\Delta\rightarrow 0$ and $C\gg1$, because the two terms in the round brackets compensate each other.
The matrix $\hat D(\boldsymbol x)$ is the approximate diffusion matrix used in the main text. 
The approximation consists in neglecting the term $d_\Delta(\boldsymbol x)$ and is justified when $\Delta \ll 1$ and the mini-batch size is large;
yet, in the experiments described in the main text, we show that the approximation continues to be useful hen $C=256$ even for $\Delta = 1$,
that is, $\alpha \in [0,2]$.

\section{Noise term in the Fokker-Planck equation}
\label{fokpla}

The continuity equation \eqref{con} for the probability density can be rewritten as%
\footnote{We leave the functional dependencies on $\boldsymbol V$, $\boldsymbol X$ and $t$ implicit throughout this appendix, except where this would lead to confusion.}
\begin{equation}
 \frac{\partial \rho}{\partial t} = 
 - \Delta_0 \rho
 - \Delta_1 \rho\ ,
\end{equation}
where we have defined the differential operators $\Delta_{0,1}$ as follows
\begin{align}
 \Delta_0 &= \boldsymbol V\, \boldsymbol \nabla_{(X)} 
 - \frac{1-\mu}{l} -\frac{1-\mu}{l}\, \boldsymbol V\, \boldsymbol \nabla_{(V)}
 - \frac{1}{l} \boldsymbol \nabla_{(X)} f \, \boldsymbol \nabla_{(V)}\ ,\\
 \Delta_1 &= \sqrt{\frac{D}{2C}} \frac{d \boldsymbol W}{dt} \, \boldsymbol \nabla_{(V)}\ .
\end{align}
We also define a ``reduced'' probability density $\tilde \rho$
\begin{equation}
 \rho= e^{-\Delta_0 t}\, \tilde \rho\ ,  
\end{equation}
so that it satisfies an equation sourced only by a term proportional to noise
\begin{equation}
 \frac{\partial \tilde \rho}{\partial t} = - e^{\Delta_0 t}\, \Delta_1\, e^{-\Delta_0 t}\, \tilde \rho\equiv -\Theta\, \tilde \rho\ ,
\end{equation}
whose solution can be expressed as
\begin{equation}\label{reduc}
 \tilde \rho(t) = e^{-\int_{t_0}^t dt' \Theta(t')} \tilde \rho(t_0)\ .
\end{equation}

We recall that the Wiener noise $\frac{d \boldsymbol W}{dt}$ is assumed to be Gaussian,
therefore also the output of the operator $\Theta$ is Gaussian. For later convenience, 
we define another Gaussian operator by integration of $\Theta$,
\begin{equation}
 I(t) = i \int_{t_0}^t dt' \Theta(t')\ .
\end{equation}
This allows us to re-express \eqref{reduc} as
\begin{equation}\label{cara}
  \tilde\rho(t) = e^{i I(t)}\, \tilde\rho (t_0) = \varphi_{I(t)}(1)\ \tilde\rho (t_0)\ ,
\end{equation}
where $\varphi_{I(t)}(1)$ is the characteristic function for $I(t)$,
\begin{equation}
 \varphi_{I(t)}(\tau) = e^{i \tau I(t)}\ ,
\end{equation}
evaluated at $\tau=1$. Relying on the Gaussianity of $I(t)$, we have
\begin{equation}
 \langle \varphi_{I(t)}(\tau)\rangle = e^{i \tau \mu_{I(t)} - \frac{1}{2}\sigma_{I(t)}^2 \tau^2}\ .
\end{equation}
Assuming that the Wiener noise is centered at zero, the mean value $\mu_{I(t)}$ vanishes.
The variance $\sigma_{I(t)}^2$ can instead be computed as follows
\begin{equation}\label{var}
 \begin{split}
 \sigma_{I(t)}^2 &= \langle I(t) I(t) \rangle = - \int_{t_0}^t dt' \int_{t_0}^t dt'' \langle \Theta(t') \Theta(t'') \rangle\\
 &= - \int_{t_0}^t dt' \int_{t_0}^t dt'' \langle e^{\Delta_0 t'}\, \Delta_1\, e^{-\Delta_0 t'} e^{\Delta_0 t''}\, \Delta_1\, e^{-\Delta_0 t''} \rangle\\
 &= - \int_{t_0}^t dt' \int_{t_0}^t dt''  \langle \boldsymbol \xi(t') \boldsymbol \xi(t'')\rangle e^{\Delta_0 t'}\, \boldsymbol \nabla_{(V)} e^{-\Delta_0 t'} e^{\Delta_0 t''}\, \boldsymbol \nabla_{(V)} e^{-\Delta_0 t''} \\
 &= - \frac{D}{C}\int_{t_0}^t dt' e^{\Delta_0 t'} \boldsymbol \nabla^2_{(V)} e^{-\Delta_0 t'}\ ,
 \end{split}
\end{equation}
where we have identified $\frac{d \boldsymbol W}{dt} = \boldsymbol \xi$ and used the noise correlation \eqref{cicova} and the assumption \eqref{iso}.
We remind the reader that \eqref{var} represents the variance of a differential operator.
We can then consider the ensemble average of \eqref{cara} obtaining
\begin{equation}\label{angolosa}
  \langle \tilde\rho(t) \rangle  = \langle e^{i I(t)}\rangle \, \tilde\rho (t_0) = \langle \varphi_{I(t)}(1)\rangle  \, \tilde\rho (t_0)
  = e^{- \frac{1}{2}\sigma_{I(t)}^2}  \, \tilde\rho (t_0)\ .
\end{equation}
Using \eqref{var} and taking the time derivative of \eqref{angolosa} we get
\begin{equation}
  \frac{\partial}{\partial t} \langle \tilde\rho(t) \rangle  = 
  \frac{D}{2C} e^{\Delta_0 t'} \boldsymbol \nabla^2_{(V)} e^{-\Delta_0 t'} \langle \tilde\rho(t) \rangle \ ,
\end{equation}
which, once expressed in terms of the original probability density, yields the Fokker-Planck equation 
\begin{equation}
  \frac{\partial}{\partial t} \langle \rho(t) \rangle  = 
  -\Delta_0 \langle \rho(t) \rangle
  + \frac{D}{2C} \boldsymbol \nabla^2_{(V)} \langle \rho(t) \rangle \ ,
\end{equation}
as in \eqref{ffpp} of the main text.

\section{Adopted neural networks}
\label{Ann}

For the experiments on MNIST we considered a shallow, bi-layer perceptron (Subsection \ref{model1}) and a deep convolutional network (Subsection \ref{model2}).
For both models we used ReLU activation functions and no dropout. The details of the architectures are given below.

For the experiments on CIFAR10 we adopted Model 3, Subsection \ref{model3}.

\subsection{Model 1: bi-layer perceptron}
\label{model1}
\begin{equation}
 \begin{array}{ccc}
  \vspace{5pt}
  \text{layer type} & \text{in channels} & \text{out channels}\\
  \text{Linear} & 28^2 & 100\\
  \text{Linear} & 100 & 10
 \end{array}
\end{equation}

\subsection{Model 2: deep convolutional network}
\label{model2}
\begin{equation}
 \begin{array}{ccc}
  \vspace{5pt}
  \text{layer type} & \text{in channels} & \text{out channels}\\
  \text{Conv2d} & 1 & 64\\
  \text{Conv2d} & 64 & 64\\
  \text{Maxpool} &  & \\
  \text{Conv2d} & 64 & 128\\
  \text{Conv2d} & 128 & 128\\
  \text{Conv2d} & 128 & 128\\
  \text{Maxpool} &  & \\
  \text{Linear} & 128*3*3 & 10\\
 \end{array}
\end{equation}

\subsection{Model 3: VGG16, deep(er) convolutional network}
\label{model3}

Model 3 coincides with the VGG16 architecture and refer to \cite{2014arXiv1409.1556S} for its description.
We used Pytorch implementation with weights pre-trained on ImageNet \cite{deng2009imagenet}.
\end{document}